\documentclass{article}
\usepackage{spconf,amsmath,graphicx,hyperref}
\usepackage{booktabs}
\usepackage{multirow}
\usepackage{makecell}
\usepackage{adjustbox}

\title{A Unified Spoken Language Model with Injected Emotional-Attribution Thinking for Human-like Interaction}
\name{
\begin{tabular}{c}
Qing Wang$^{\dag}$, Zehan Li$^{\dag}$, Yaodong Song$^{\dag}$, Hongjie Chen$^{\dag}$\thanks{$^{\dag}$ Equal contribution.}, Jian Kang, \\ Jie Lian, Jie Li, Yongxiang Li, Xuelong Li$^*$\thanks{$*$ Corresponding author. Email: xuelong\_li@ieee.org}
\end{tabular}
}
\address{Institute of Artificial Intelligence (TeleAI), China Telecom, China}
\begin{document}
\ninept
\maketitle
\begin{abstract}

This paper presents a unified spoken language model for emotional intelligence, enhanced by a novel data construction strategy termed Injected Emotional-Attribution Thinking (IEAT). IEAT incorporates user emotional states and their underlying causes into the model’s internal reasoning process, enabling emotion-aware reasoning to be internalized rather than treated as explicit supervision. The model is trained with a two-stage progressive strategy. The first stage performs speech–text alignment and emotional attribute modeling via self-distillation, while the second stage conducts end-to-end cross-modal joint optimization to ensure consistency between textual and spoken emotional expressions. Experiments on the Human-like Spoken Dialogue Systems Challenge (HumDial) Emotional Intelligence benchmark demonstrate that the proposed approach achieves top-ranked performance across emotional trajectory modeling, emotional reasoning, and empathetic response generation under both LLM-based and human evaluations.

\end{abstract}
\section{Introduction}
\label{sec:intro}

Spoken dialogue systems are evolving beyond task-oriented interactions toward emotional intelligence, requiring models to understand and respond to dynamic human emotions. However, achieving semantic and emotional congruence within a unified speech-language framework remains a significant challenge. To address this, the the Human-like Spoken Dialogue Systems (HumDial) Challenge\footnote{\url{https://github.com/ASLP-lab/Hum-Dial/tree/main}} – Track 1: Emotional Intelligence focuses on pushing the boundaries of emotional reasoning and response generation in multimodal systems.

\begin{figure}
    \centering
    \includegraphics[width=1\linewidth]{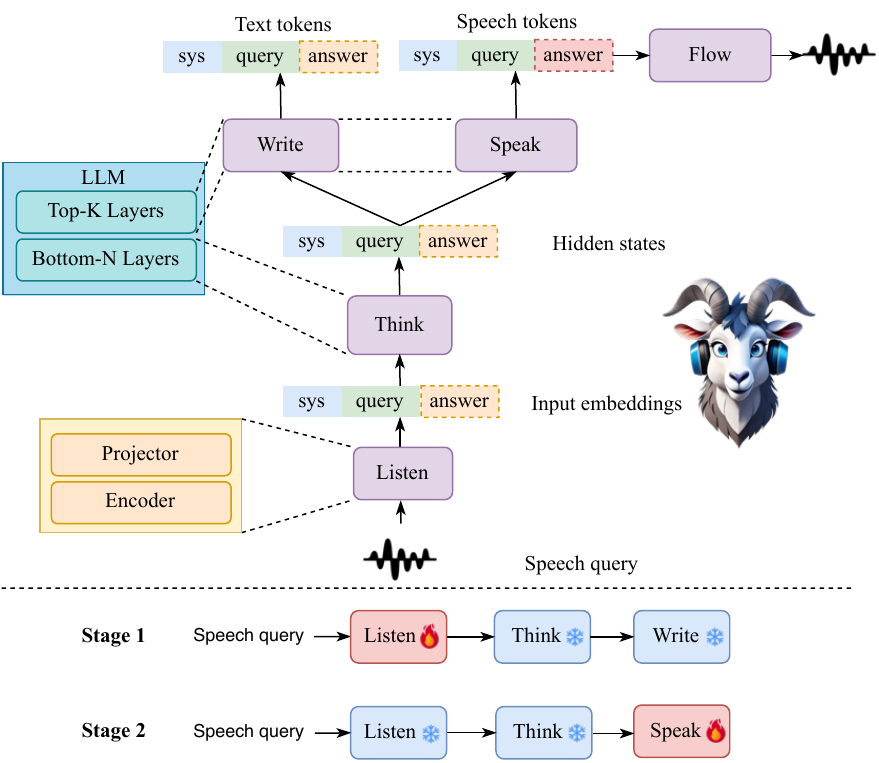}
    \caption{Overview of training procedure.}
    \label{fig:slm}
\end{figure}

In this study, our system is built upon the GOAT-SLM~\cite{chen2025goat} architecture, which is implemented within the AI Flow framework~\cite{an2506ai} and incorporates affective cues from BoSS~\cite{wang2025boss}. It follows a Listen–Think–Write–Speak paradigm to process speech and text in a shared semantic space and supporting coherent generation of both text and speech outputs.
To enhance emotional intelligence, we propose Injected Emotional-Attribution Thinking (IEAT), a data construction and training strategy that explicitly integrates emotional states and their inducing factors into the model’s internal thinking process.
Our approach ranked first in the HumDial Emotional Intelligence track, demonstrating superior and consistent performance across all three subtasks under the challenge’s comprehensive evaluation protocol. These results highlight the potential of unified speech–language modeling augmented with emotional reasoning injection as a promising direction for emotionally intelligent spoken dialogue systems.

\begin{table*}[ht]
\centering
\begin{adjustbox}{max width=1\textwidth}
\renewcommand{\tabcolsep}{0.13cm}
\renewcommand\arraystretch{1.0}
\begin{tabular}{lcccccccc}
\toprule
\textbf{Model} & \textbf{Task1-zh} & \textbf{Task1-en} & \textbf{Task2-zh} & \textbf{Task2-en} & \textbf{Task3-zh} & \textbf{Task3-en} & \textbf{AQA-zh~\cite{li2025televal}} & \textbf{AQA-en~\cite{li2025televal}} \\
\midrule
Freeze-omni~\cite{wang2024freeze} & 3.11 & 2.58 & 3.63 & 2.79 & 4.02 & 3.66 & 21.23 & 45.29 \\
Stepaudio2-mini~\cite{wu2025step} & 3.71 & 2.22 & 3.81 & 3.16 & \underline{4.53} & 4.30 & 28.11 & 47.67 \\
Qwen2.5-omni~\cite{xu2025qwen2} & 4.38 & \underline{4.40} & 3.25 & 3.73 & 4.28 & 4.18 & 22.08 & 50.13 \\
Qwen3-omni~\cite{xu2025qwen3} & 4.05 & 3.90 & 3.80 & 3.56 & 4.34 & \underline{4.31} & \textbf{49.95} & \underline{56.91} \\
Mimo-audio-instruct~\cite{coreteam2025mimoaudio} & \underline{4.89} & 3.92 & \underline{4.89} & \underline{4.51} & 4.00 & 3.63 & 34.29 & 51.63  \\
\midrule
\textbf{Our Model} & \textbf{4.98} & \textbf{4.87} & \textbf{4.98} & \textbf{4.83} & \textbf{4.53} & \textbf{4.36} & \underline{37.38} & \textbf{57.69} \\
\bottomrule
\end{tabular}
\end{adjustbox}
\caption{Results on dev set of HumDial Challenge Track 1. Task* scores follow the official LLM-based evaluation (0-5 scale $\uparrow$). AQA* scores follow the TELEVAL (0-100 scale $\uparrow$). The highest score is in bold and the second highest is underlined.}
\label{tab:emotion_results}
\end{table*}

\section{System Description}
\label{sec:format}

\subsection{System Overview}

Our system adopts an end-to-end architecture that integrates speech perception, semantic reasoning, and multimodal generation. As shown in Figure~\ref{fig:slm}, the model is grounded in the GOAT-SLM structure, where speech inputs are mapped into a shared semantic space and processed by a common reasoning core to produce consistent text and speech outputs.

Training is performed using a two-stage progressive strategy. The first stage focuses on speech–text alignment and emotional attribute modeling through self-distillation, gradually extending from semantic alignment to the joint modeling of linguistic content, paralinguistic information, and speaker characteristics. The second stage conducts end-to-end cross-modal joint optimization, simultaneously training text and speech generation pathways to improve emotional coherence and semantic consistency across modalities. The speech-generation branch employs modular fine-tuning on top-k layers of an LLM for speech token prediction while freezing the bottom-n layers to preserve foundational linguistic knowledge. Moreover, multi-token prediction is introduced to support real-time streaming TTS synthesis.


To further enhance emotional understanding, we introduce Injected Emotional-Attribution Thinking (IEAT) as a data construction strategy. IEAT injects emotional states and their underlying causes into the model’s internal reasoning process, allowing emotional information to be internalized as part of the model’s cognition. Together, the unified architecture, two-stage training, and IEAT enable effective emotional trajectory modeling, emotional reasoning, and empathetic response generation within a single spoken language model.

\subsection{Data Construction}


We employ the Qwen3-8B model to generate both training targets and corresponding instructions. The training set consists of Chinese–English subsets drawn from the open-source Emilia~\cite{he2024emilia}, WenetSpeech4TTS~\cite{ma2024wenetspeech4tts}, and GigaSpeech~\cite{chen2021gigaspeech} datasets. We additionally sample suitable user queries from train\_3.5M\_CN~\cite{ji2023better,BELLE} and Ultrachat~\cite{ding2023enhancing}, and use CosyVoice2~\cite{du2024cosyvoice} model to synthesize speech queries. Then, we merge these samples into the training set. In stage 1-1, we feed either the textual queries or ASR transcripts into the LLM to produce natural responses, which then serve as supervision for cross-modal alignment between the audio and text modalities. In Stage 1-2, to encourage the model to extract emotional cues from acoustic information rather than relying solely on textual semantics, we apply the emo2vec~\cite{xu2018emo2vec} model to perform emotion classification on all Stage 1-1 audio samples, obtaining emotion labels for each example. We then emulate the analytical style used by Qwen3-8B in its internal thinking-mode by inserting the emotion information into the think body, omitting the think-termination token (\verb|<\think>|) so that Qwen3-8B continues generating both its reasoning trace and the final response. This procedure constitutes the IEAT step. Ultimately, the outputs for stage 1-2 are obtained by applying prompts corresponding to the three tasks. To strengthen the model’s multi-turn dialogue capabilities and its performance on the emotional-trajectory detection task, we sample 40\% of the constructed empathetic-response training data and reorganize it into 2-turn, 3-turn, and 4-turn dialogues, ensuring that the initial and final emotional states differ.

In addition, we create instruction audios for stage 1-2. Using LLM generation followed by manual filtering, we produce 500 text instructions for each task, and synthesize with CosyVoice2 and 50 randomly speakers selected from WenetSpeech4TTS. These instruction audios are then concatenated with the stage 1-2 training audio. The empathetic-response task does not require extra instruction audio. In Stage 2, we train on Emilia and LibriHeavy~\cite{kang2023libriheavy} to further strengthen cross-modal alignment and speech synthesis quality.



\section{Experimental Results}
\label{sec:exp}

Under the TELEVAL framework~\cite{li2025televal}, we evaluate our SLM on the official development set of the HumDial and audio question answering (AQA) sets in TELEVAL. This allows for a rigorous comparison against several strong open-source baselines, including freeze-omni~\cite{wang2024freeze}, stepaudio2-mini~\cite{wu2025step}, qwen2.5-omni~\cite{xu2025qwen2}, Qwen3-omni~\cite{xu2025qwen3}, and Mimo-audio-instruct~\cite{coreteam2025mimoaudio}. 
SLMs inference with a default system prompts and all scores follow the challenge’s official LLM-based scoring guidelines. 
As shown in Table~\ref{tab:emotion_results}, the results show that our SLM achieves state-of-the-art emotional intelligence across multilingual emotional trajectory detection, emotional reasoning, and empathetic response. The consistent performance gains across tasks and languages validate the effectiveness of our unified speech–language modeling approach, while keeping the intelligence of common sense.

\begin{table}[ht]
\centering
\begin{adjustbox}{max width=1\textwidth}
\renewcommand{\tabcolsep}{0.2cm}
\renewcommand\arraystretch{1.0}
\begin{tabular}{lcccccc}
\toprule
\textbf{Team} & \textbf{Task1} & \textbf{Task2} & \textbf{Task3} & \textbf{Hum} & \textbf{Final Score} \\
\midrule
\textbf{Ours (1st)} & 4.97 & 4.98 & 3.85 & 3.79 & 4.27 \\
Rank 2nd & 4.90 & 5.00 & 4.14 & 3.70 & 4.24 \\
Rank 3rd & 4.76 & 4.76 & 4.02 & 3.81 & 4.21 \\
Rank 4th & 3.67 & 4.92 & 4.93 & 3.71 & 4.06 \\
\midrule
Baseline & 2.62 & 2.73 & 2.73 & 2.96 & 2.82\\
\bottomrule
\end{tabular}
\end{adjustbox}
\caption{Results on test set of HumDial Challenge Track 1. Scores follow the official LLM-based evaluation (0–5 scale $\uparrow$). }
\label{tab:emotion_results3}
\end{table}

The final results on the test set are shown in Table~\ref{tab:emotion_results3}, our model achieves the highest final score of 4.27, ranking first among all participating systems. Although some competing methods obtain comparable or slightly higher scores on individual subtasks, our approach consistently maintains strong performance across Task 1, Task 2, and the human evaluation dimension, leading to the best overall weighted score, highlighting the effectiveness of the proposed unified spoken language modeling with IEAT strategy.

\section{Conclusion}
In this study, we introduce a unified spoken language model for emotional intelligence, enhanced by Injected Emotional-Attribution Thinking (IEAT). By internalizing emotional states and their causes into the model’s reasoning process, the proposed approach effectively supports emotional intelligence within a single framework. Evaluated on the HumDial Emotional Intelligence benchmark, our method ranks first in the competition, demonstrating the effectiveness of unified spoken language modeling with emotional reasoning injection.

\bibliographystyle{IEEEbib}
\bibliography{strings,refs}

\end{document}